\documentclass[runningheads]{llncs}

\usepackage{eccv}
\usepackage{eccvabbrv}

\usepackage{graphicx}
\usepackage{booktabs}
\usepackage{pifont}
\usepackage{multirow}   % 提供 \multirow 功能
\usepackage{xcolor}     % 提供 \textcolor 功能
\usepackage{makecell} % 必须添加此包用于换行
\usepackage{tabularx} % 用于控制表格总宽度
\usepackage{pifont}
\usepackage{siunitx}  % 用于 S 列类型，实现数字对齐

\usepackage[accsupp]{axessibility}  % Improves PDF readability for those with disabilities.

\usepackage{hyperref}

\usepackage{orcidlink}

\begin{document}

\title{Sketch and Text Synergy: Fusing Structural Contours and Descriptive Attributes for Fine-Grained Image Retrieval} 

\titlerunning{Sketch and Text Synergy for Fine-Grained Image Retrieval}

\author{Siyuan Wang\inst{1} \and
Hanchen Gao\inst{1} \and
Guangming Zhu\inst{1} \and
Jiang Lu\inst{1} \and
Yiyue Ma\inst{2} \and
Tianci Wu\inst{1} \and
Jincai Huang\inst{1} \and
Liang Zhang \inst{1}
}

\authorrunning{F.~Author et al.}

\institute{Xidian University \and First Aircraft Design Institute}

\maketitle

\begin{abstract}
Fine-grained image retrieval via hand-drawn sketches or textual descriptions remains a critical challenge due to inherent modality gaps. While hand-drawn sketches capture complex structural contours, they lack color and texture, which text effectively provides despite omitting spatial contours. Motivated by the complementary nature of these modalities, we propose the Sketch and Text Based Image Retrieval (STBIR) framework. By synergizing the rich color and texture cues from text with the structural outlines provided by sketches, STBIR achieves superior fine-grained retrieval performance.
First, a curriculum learning driven robustness enhancement module is proposed to enhance the model’s robustness when handling queries of varying quality. 
Second, we introduce a category-knowledge-based feature space optimization module, thereby significantly boosting the model's representational power. 
Finally, we design a multi-stage cross-modal feature alignment mechanism to effectively mitigate the challenges of cross modal feature alignment.
Furthermore, we curate the fine-grained STBIR benchmark dataset to rigorously validate the efficacy of our proposed framework and to provide data support as a reference for subsequent related research. Extensive experiments demonstrate that the proposed STBIR framework significantly outperforms state of the art methods.
\end{abstract}

\section{Introduction}
\label{sec:intro}

\begin{figure}[t] % h:当前位置, t:顶部, b:底部, p:独立一页
	\centering
	\includegraphics[width=\textwidth]{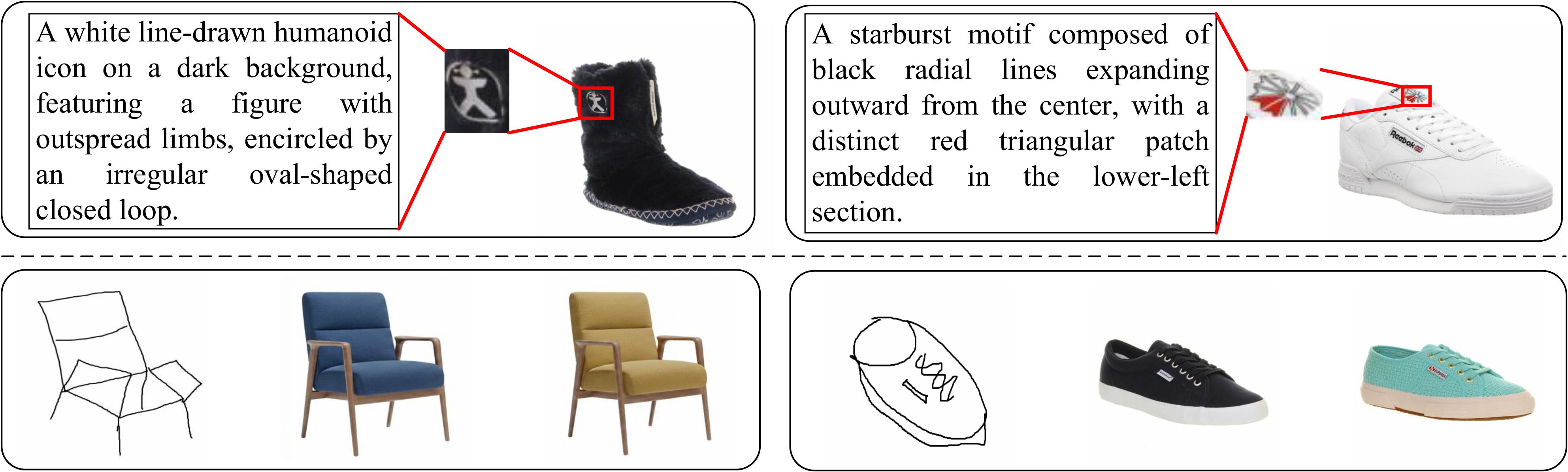} 
	\caption{Characteristics of different query modalities in fine-grained retrieval. Row 1 illustrates that textual descriptions often struggle to accurately convey irregular shapes and complex spatial structures. Row 2 demonstrates that when target instances differ primarily in color, hand-drawn sketches fail to provide sufficient discriminative cues for accurate retrieval.}
	\label{fig:Modality_description} % 用于文中引用，如：见图 \ref{fig:my_label}
\end{figure}

Users can intuitively express target objects through either textual descriptions or hand-drawn sketches, making both modalities prevalent in fine-grained image retrieval (FG-IR) \cite{chowdhury2023scenetrilogy,sangkloy2022sketch}. At the semantic level, while text excels at characterizing visual attributes such as color and texture, it struggles to convey intricate structural details (e.g., the irregular ornaments in \cref{fig:Modality_description}, Row 1). Conversely, sketches are adept at delineating geometric contours but inherently lack chromatic and textural information. In fine-grained scenarios, where inter-instance variances frequently involve non geometric attributes such as color (e.g., \cref{fig:Modality_description}, Row 2), relying solely on the sketch modality limits overall retrieval performance. Consequently, text and sketches offer a potent semantic complementarity, enabling a significantly more comprehensive representation for FG-IR.

Previous works have established the intrinsic semantic complementarity between text and hand-drawn sketches, prompting research into multimodal fusion for image retrieval tasks. For instance, the SBCIR framework proposed by Koley et al. \cite{koley2024you} leverages generated pseudo textual descriptions to complement initial sketches, effectively mitigating the challenges associated with sparse fine-grained annotations. Consequently, Gatti et al. \cite{gatti2024composite} introduced STNet, exploiting user sketches to successfully localize subjects across naturally occurring scenes, facilitating robust retrieval through joint text image encoding. Nevertheless, these studies have not sufficiently addressed the profound distribution mismatch in cross-modal feature spaces, a problem inherently driven by the modality gap. This unaddressed disparity frequently results in suboptimal feature alignment and exacerbates parameter divergence, causing the convergence instability frequently observed during joint representation learning. To tackle the aforementioned challenges, we propose the STBIR framework, which aims to construct a robust cross-modal feature space, thereby achieving superior fine-grained retrieval performance.

The STBIR framework comprises three pivotal modules. 
First, we introduce a robustness module that progressively injects noise to simulate perturbations, thereby strengthening the model against low-quality queries.
Second, to enhance instance-level discriminability, we design a category-knowledge-guided optimization module. This module integrates category priors as constraints to refine the feature representation space.
Finally, to mitigate parameter divergence arising from inconsistent gradient magnitudes during multimodal joint optimization, we propose a multi-stage cross-modal feature alignment mechanism, which effectively prevents feature space perturbations and enhances overall retrieval stability.  

Addressing the scarcity of datasets for tri-modal fine-grained image retrieval, we construct the STBIR dataset, which comprises three subsets: STBIR-S, STBIR-C, and STBIR-D. Specifically, STBIR-S and STBIR-C focus on fine-grained retrieval for single categories (shoes and chairs, respectively), while STBIR-D encompasses fine-grained retrieval across a large scale of daily object categories. 

In summary, our main contributions are as follows:

\begin{itemize}
	\item We introduce the STBIR framework, which integrates textual attribute descriptions with sketched structural information. This holistic modeling of fine-grained features significantly enhances fine-grained retrieval performance.
	\item We propose a curriculum learning-driven robustness enhancement module and a category-knowledge-based feature space optimization Module, significantly bolstering the model's discriminability and robustness.
	\item We propose a multi-stage cross-modal feature alignment mechanism, which effectively optimizes the feature space and enhances the representation capability of tri-modal data.
	\item We create and release a dedicated benchmark for tri-modal fine-grained image retrieval. This benchmark provides a valuable resource for the community to explore sketch-and-text-based image retrieval.
\end{itemize}

\section{Related Work}
\subsection{Fine-grained Sketch Based Image Retrieval (FG-SBIR)}
Fine-grained sketch based image retrieval necessitates models to forge precise correspondence between hand-drawn sketches and image outputs at the instance-level \cite{bhunia2021more,sun2022dli}. Deep dual siamese network architectures optimized via triplet loss represent the predominant framework within this domain \cite{yu2016sketch,sangkloy2016sketchy}. To enhance model discriminability regarding detailed information, researchers have employed generative tasks to enable encoders to retain more fine-grained details \cite{pang2017cross}, or introduced soft attention mechanisms to capture highly discriminative features \cite{song2017deep}. In response to the scarcity of paired fine-grained data, strategies such as semi-supervised learning \cite{bhunia2021more} and self-supervised pretraining \cite{pang2020solving} have been successively proposed. Furthermore, addressing the inherent abstraction of sketches and variances in user drawing, some work has extensively explored topics including occluded sketches \cite{bhunia2020sketch}, noise robustness \cite{bhunia2022sketching}, hierarchical abstraction modeling \cite{sain2020cross}, and the diversity of drawing styles \cite{sain2021stylemeup,bhunia2022adaptive}.

\subsection{Fine-grained Text based image retrieval (FG-TBIR)}
Fine-grained Text Based Image Retrieval aims to retrieve semantically matching image instances from large scale databases using natural language descriptions. While early global feature alignment approaches focused on holistic semantic mapping \cite{faghri2017vse++,wang2018learning}, their performance was often constrained by a neglect of detailed visual attributes described in text. To overcome this bottleneck, local alignment mechanisms have emerged as the primary research focus. By leveraging cross attention \cite{lee2018stacked}, models establish point to point correlations between image regions and textual tokens, facilitating precise feature alignment. Building on this, recent studies have explored multi-level and relation aware alignment strategies \cite{chen2020imram, zhang2020context , ji2021step , xue2021probing , yu2021ernie }; this transition from single target matching to structured relational alignment significantly enhances the capability to handle complex, detail rich descriptions, laying a solid foundation for more intelligent retrieval systems.

\subsection{Fine-grained Sketch and Text Based Image Retrieval (FG-STBIR)}
\label{related_work:fgstbir}
Sketch and text based image retrieval achieves instance level retrieval by fusing the structural information of sketches with the color and texture attributes of text. Currently, this field faces dual challenges: data scarcity and difficulties in aligning multimodal feature spaces. To mitigate data absence, one prominent approach converts images into sketch styles via edge detection, yet this strategy leads to severe loss of color and depth features \cite{song2017fine}. Alternatively, other work employs conditional Invertible Neural Networks (cINN) to disentangle modality specific and agnostic features, enabling flexible joint embeddings for diverse retrieval and generation tasks \cite{chowdhury2023scenetrilogy}. With the rise of large vision-language models, establishing tri-modal alignment among "sketch-text-photo" via CLIP has emerged as a new paradigm \cite{sangkloy2022sketch,chowdhury2023scenetrilogy}. While these methods effectively transfer representation knowledge from large models to aid feature extraction, achieving precise feature space alignment for fine-grained tasks remains a core-challenge.

%With the advent of Vision-Language Models (VLMs), leveraging CLIP to establish a tri-modal alignment (sketch-text-photo) has emerged as a new paradigm \cite{sangkloy2022sketch,chowdhury2023scenetrilogy}. Although these methods effectively transfer foundational representation knowledge to facilitate feature extraction, achieving precise feature-space calibration specifically for fine-grained tasks remains a formidable hurdle.

%However, distinct modalities exhibit significant discrepancies in convergence rates and gradient contributions during the training process. Dominant modalities with superior performance tend to generate larger gradient magnitudes, thereby monopolizing the optimization process. This phenomenon often leads to the under-optimization of other modalities, or worse, triggers model oscillations and parameter divergence \cite{wu2022characterizing, peng2022balanced, liang2022mind}. To address these challenges, we propose the STBIR framework. By refining the feature-space alignment mechanism across modalities, our approach effectively balances the multi-modal learning process, yielding substantial improvements in retrieval performance.

\subsection{Dataset}
Currently, benchmarks that simultaneously integrate hand-drawn sketches, textual descriptions, and natural images are notably scarce. While TSBIR \cite{sangkloy2022sketch} introduces a tri-modal setting, its training sketches are synthesized via method \cite{li2019photo} rather than being human authored. Song et al. \cite{song2017fine} curate a fine-grained dataset of 1,112 shoe instances, yet it remains inaccessible to the public. In the CSTBIR dataset \cite{gatti2024composite}, although images and texts are paired at the instance-level, sketches sourced from QuickDraw maintain only category-level alignment with the images. Similarly, SketchyCOCO \cite{gao2020sketchycoco} provides tri-modal data, but its sketches are derived from predefined templates rather than being authentic, instance-specific tracings of the corresponding objects. 

\section{Dataset}
\label{sec:dataset}
We introduce the STBIR dataset, a tri-modal fine-grained retrieval benchmark designed for two distinct scenarios: single-category retrieval (encompassing STBIR-S and STBIR-C) and retrieval across extensive daily categories (STBIR-D).

%\begin{figure}[t] % h:当前位置, t:顶部, b:底部, p:独立一页
%	\centering
%	\includegraphics[width=\textwidth]{images/data_comparsion.pdf} 
%	\caption{Comparison of mainstream SBIR datasets. In CSTBIR, sketches and images maintain only category level correspondence. In TSBIR, images are primarily sourced from the scene oriented COCO dataset with algorithmically generated training sketches. In contrast, QMUL Shoe, QMUL Chair, and Sketchy strictly adhere to manual drawing, ensuring high instance level consistency between images and sketches.}
%	\label{fig:stbir_dataset_comparsion} % 用于文中引用，如：见图 
%\end{figure}

\subsection{Visual Data Sources and Characteristics}
Building upon the visual data from the QMUL-Shoe, QMUL-Chair, and Sketchy datasets, we construct the STBIR tri-modal fine-grained retrieval benchmark by enriching these images with detailed textual descriptions. As illustrated in \cref{fig:qmul_sketchy_text_dataset}, the STBIR-S and STBIR-C subsets focus on single-category objects with backgrounds removed, making them ideal for vertical scenarios such as product search. In contrast, the STBIR-D subset encompasses diverse daily objects embedded in rich contextual backgrounds, thereby catering to broader and more complex retrieval demands.

\subsection{LLM-based Textual Description Generation and Manual Verification}
We leverage the multimodal large language model Qwen\footnote{https://www.qianwen.com} to generate structured descriptions of image subjects by designing a schema-guided prompting strategy: 1) focusing on the primary object [\textit{CLASS}]; 2) encompassing chromatic and textural attributes, such as leather or wooden; 3) optionally incorporating salient geometric features, such as curved back or tufted cushion; 4) restricting the output to three to twelve phrases; and 5) ensuring that phrases are comma separated without additional punctuation or full sentences. 

\begin{figure}[t] % h:当前位置, t:顶部, b:底部, p:独立一页
	\centering
	\includegraphics[width=\textwidth]{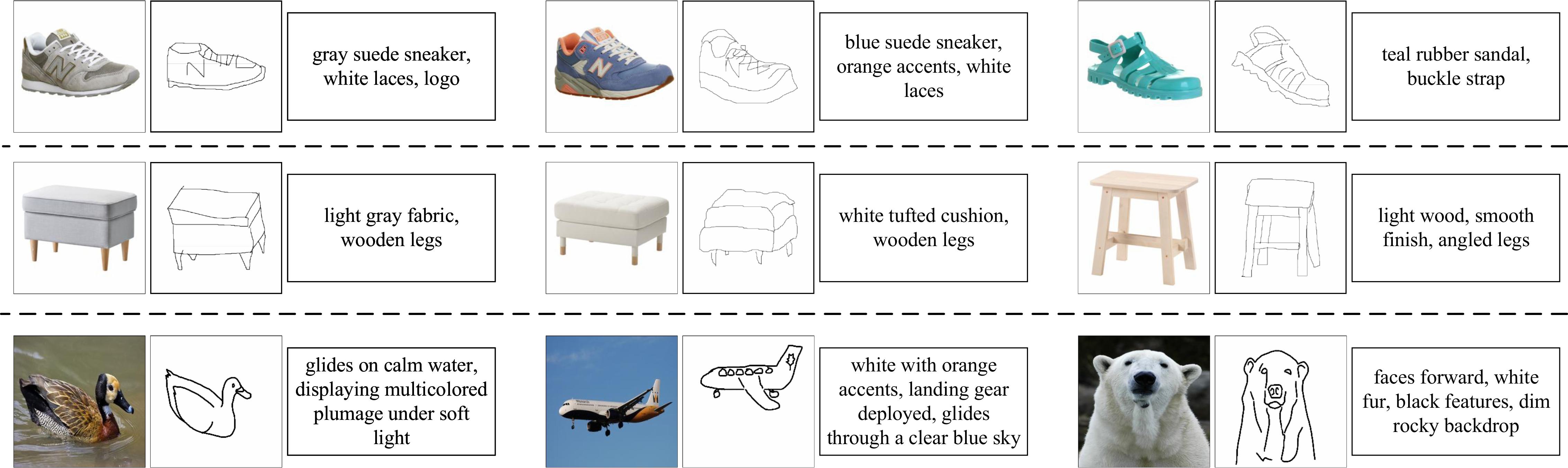} 
	\caption{Visualization of samples from the STBIR dataset. Row 1 shows instances from the STBIR-S subset. Row 2 displays examples from STBIR-C.  Row 3 presents samples from STBIR-D. For each sample, the sketch, text, and image are strictly aligned.}
	\label{fig:qmul_sketchy_text_dataset} % 用于文中引用，如：见图 
\end{figure}

The raw generated descriptions subsequently undergo manual verification to eliminate entries with semantic drift or redundant descriptions, thereby ensuring high-fidelity alignment and accuracy of the tri-modal paired data across fine-grained attributes. \cref{fig:qmul_sketchy_text_dataset} presents some instances of the proposed dataset.

\subsection{Semantic Augmentation and Comparative Analysis}
To enhance the semantic richness of the datasets, this study performs semantic category annotation on the original data. For the QMUL-Shoe dataset, which initially contained only a single "shoe" macro-category, we subdivid it into 13 granular subcategories, such as sneakers and slippers. Similarly, instances in QMUL-Chair are annotated with 24 fine-grained category labels. In contrast, for the Sketchy dataset, which already possesses a relatively rich predefined category hierarchy, we preserve the original semantic categorization.

\cref{tab:dataset_comparison} presents a comparison of the characteristics of datasets containing sketches, text, and images. As observed, the proposed dataset is the only one that offers fine-grained matching across all three modalities, features hand-drawn sketches, and is publicly available.

\begin{table}[htbp]
	\centering
	\caption{Comparison of tri-modal datasets for sketch and text based image retrieval. The "Hand-drawn" indicates whether the sketches are exclusively manually drawn. The TSBIR dataset does not provide category classifications (marked as "/").}
	\label{tab:dataset_comparison}
	% 计算比例：总共6列。令普通列宽为x，第一列为2x。总宽度=7x。
	% 第一列占 2/7，分配的 \hsize = (2/7) * 6 = 1.714
	% 其余列占 1/7，分配的 \hsize = (1/7) * 6 = 0.857
	\begin{tabularx}{\textwidth}{
			>{\hsize=1.714\hsize\centering\arraybackslash}X
			>{\hsize=0.857\hsize\centering\arraybackslash}X
			>{\hsize=0.857\hsize\centering\arraybackslash}X
			>{\hsize=0.857\hsize\centering\arraybackslash}X
			>{\hsize=0.857\hsize\centering\arraybackslash}X
			>{\hsize=0.857\hsize\centering\arraybackslash}X
		}
		\toprule
		Dataset & \makecell{Fine \\ grained} & \makecell{Hand \\ drawn} & public &\makecell{Sample \\ amount} & category \\
		\midrule
		TSBIR\cite{sangkloy2022sketch}           & \checkmark & $\times$   & \checkmark & 5000  & /   \\
		DSSA\cite{song2017fine}            & \checkmark & \checkmark & $\times$   & 1112  & 1   \\
		CSTBIR\cite{gatti2024composite}          & $\times$   & \checkmark & \checkmark & 2M    & 258 \\
		SketchyCOCO\cite{gao2020sketchycoco}     & $\times$   & \checkmark & \checkmark & 20000 & 14  \\
		STBIR-S  & \checkmark & \checkmark & \checkmark & 2000  & 13  \\
		STBIR-C & \checkmark & \checkmark & \checkmark & 400   & 24  \\
		STBIR-D    & \checkmark & \checkmark & \checkmark & 12500 & 125 \\
		\bottomrule
	\end{tabularx}
\end{table}

%\begin{table}[h]
%	\centering
%	\caption{Statistical Analysis of the Proposed Datasets}
%	\label{tab:stbir_proposed_dataset}
	% 使用 0.8\textwidth 设置总宽度，并通过 @{\extracolsep{\fill}} 自动分配列间距
%	\begin{tabular*}{0.8\textwidth}{@{\extracolsep{\fill}}lcccc}
%		\toprule
%		\textbf{Dataset} & \textbf{Samples} & \textbf{Train} & \textbf{Test} & \textbf{Category}\\ \midrule
%		QMUL-Shoe-Text  & 2000  & 1800  & 200   & 13      \\
%		QMUL-Chair-Text & 400    & 300    & 100   & 24      \\
%		Sketchy-Text    & 12500 & 11250 & 1250 & 125  \\ \bottomrule
%	\end{tabular*}
%\end{table}

\section{Methodology}
To address the critical challenges of unstable joint optimization, insufficient feature discriminability, and limited model robustness in tri-modal fine-grained retrieval, we propose the STBIR framework.
First, to enhance the retrieval robustness against low-quality queries, we propose a curriculum learning-driven robustness enhancement module. 
Second, to address the stringent demands for feature discriminability in fine-grained retrieval, we design a category-knowledge-based feature space optimization module.
Finally, to tackle the prevalent issues of gradient imbalance and parameter divergence in multimodal joint optimization, we introduce a multi-stage cross-modal feature alignment mechanism.

\begin{figure}[t] % h:当前位置, t:顶部, b:底部, p:独立一页
	\centering
	\includegraphics[width=\textwidth]{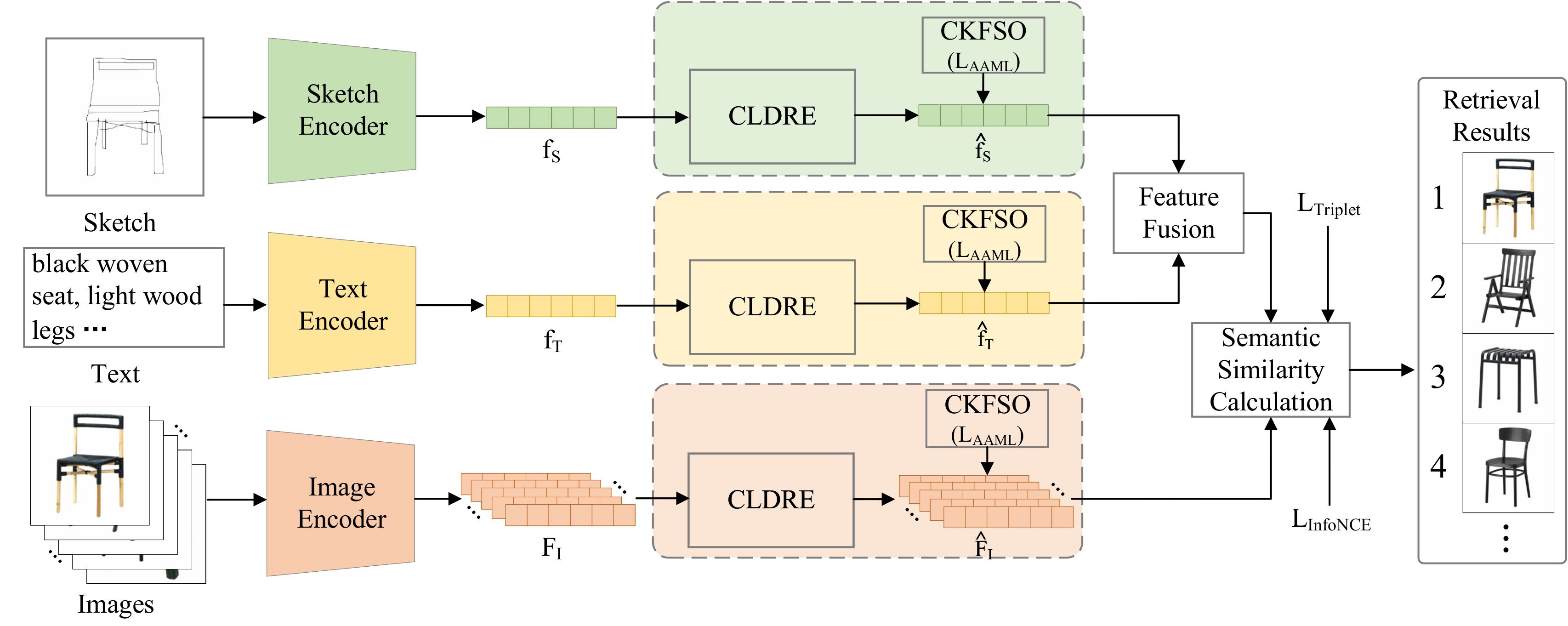} 
	\caption{Illustration of the STBIR framework. CLDRE denotes the Curriculum Learning Driven Robustness Enhancement module, while CKFSO represents the Category-Knowledge-Based Feature Space Optimization module.}
	\label{fig:stbir_network} % 用于文中引用，如：见图 
\end{figure}

\subsection{Problem Formulation}
Let $D = \{(S_i, T_i, I_i)\}_{i=1}^N$ denote a collection of $N$ paired instances, where each triplet includes: a hand-drawn sketch $S_i$ representing the geometric contours and topological layout of the target, a complementary textual description $T_i$ characterizing the chromatic, textural, and qualitative attributes of the object, and a corresponding natural image $I_i$. The proposed STBIR framework first extracts hand-drawn sketch features $f_S$ and text features $f_T$, subsequently fusing these two representations. Concurrently, it extracts the features $F_I$  of candidate images from the image gallery. Finally, by calculating the semantic similarity between the fused features $f_c$ and the candidate image features $F_I$ within the gallery, the framework retrieves the target image exhibiting the highest matching score.

\subsection{Feature Encoder}
\textbf{Hand-drawn sketch feature encoder.} Hand-drawn sketches are typically characterized in various formats, including raster images, dynamic stroke sequences, and structured graph representations. Following study \cite{xu2022deep}, convolutional neural networks exhibit inherent advantages in extracting global features and can more effectively align with the image modality within a unified feature space. We employ ResNet \cite{he2016deep} to extract features from hand-drawn sketches.

\textbf{Text and image feature encoder.} The representation capabilities acquired by CLIP\cite{radford2021learning} through large scale multi-modal data enable it to establish associations between visual images and textual semantics within a unified latent embedding space. We employ CLIP as the feature encoder for both image and text modalities. 

\subsection{Curriculum Learning Driven Robustness Enhancement (CLDRE)}
To bolster the model's fine-grained retrieval capabilities amidst low-quality inputs, we introduce a Curriculum Learning-Driven Robustness Enhancement \\{(CLDRE)} module. Grounded in the principle of simulating a cognitive evolution from simple to complex scenarios, this module systematically fortifies the framework's discriminative robustness against degraded data.

The input feature $f$ for the CLDRE module corresponds to the representations $f_S$, $f_T$, or $F_I$ extracted by the feature encoders. During the training process, we dynamically introduce noise to the input feature $f$ via \cref{eq:CLDRE} to artificially degrade the query quality, thereby enhancing the model robustness against low quality queries:

\begin{equation}
	\label{eq:CLDRE}
	\hat{f} =  \mathtt{CLDRE}(f) = f + t \cdot \alpha
\end{equation}
Here, $t \in [0,1]$ denotes the normalized training progress, and $\alpha$ represents the target noise intensity. During the initial training phase, the model primarily processes high fidelity and complete modal samples, establishing a fundamental mapping of the cross modal feature space through stable data distributions. As the training iteration $t$ increases, the perturbation level of the input data is gradually elevated.

\subsubsection{Category-Knowledge-Based Feature Space Optimization (CKFSO).}
Leveraging critical semantic priors from retrieval target categories to optimize the network significantly bolsters feature representation. To this end, we propose a category-knowledge-based feature space optimization module,  grounded in Additive Angular Margin Loss~\cite{deng2019arcface}. This approach effectively compacts intra-class distributions while expanding inter-class margins, thereby substantially elevating the overall quality of model representations.

The CKFSO module takes the augmented feature $\hat{f}$ derived from the CLDRE module and its corresponding semantic category label as inputs. By optimizing the objective function formulated in \cref{eq:stbir_arc_face}, the model minimizes the angular distance $\theta_i$ between the feature $\hat{f_i}$ and its ground-truth category feature center $f_{cc_i}$ while maximizing inter-class separability. Specifically, introducing a fixed additive angular margin penalty $m$ enforces strict intra-class feature cohesion and significantly amplifies the discriminability among features belonging to different categories. Ultimately, this margin-based optimization significantly elevates the overall representation capability of the model.

\begin{equation}
	\label{eq:stbir_arc_face}
	L_{AAML} = -\frac{1}{N} \sum_{i=1}^{N} \log \frac{e^{s \cdot \cos(\theta_{i} + m)}}{e^{s \cdot \cos(\theta_{i} + m)} + \sum_{j=1, j \neq i}^{C} e^{s \cdot \cos \theta_{j}}}
\end{equation}

\begin{equation}
	\label{eq:arcos}
	\theta_{i} = \arccos \left( \frac{\hat{f}_i \cdot f_{cc_{i}}}{\|\hat{f}_i\| \|f_{cc_{i}}\|} \right)
\end{equation}
where $N$ denotes the number of samples in a batch; $s$ represents the scaling factor that ensures the stability of the convergence process; $C$ is the total number of categories; and $f_{cc_i}$ is parameterized as the corresponding weight vector in the classification layer.

\subsection{Multi-stage Cross-modal Feature Alignment (MCFA)}
Existing research \cite{wu2022characterizing, peng2022balanced, liang2022mind} indicates that the primary cause of parameter divergence in multi-modal learning stems from distribution discrepancies between different modalities within the feature space. During the initialization phase, neural networks map images and text into mutually non-overlapping subspaces, resulting in a significant modality gap. This spatial isolation leads to drastic fluctuations in gradient computation during contrastive learning. Simultaneously, constrained by the greedy optimization nature of deep learning models, networks tend to prioritize fitting dominant modalities with more easily extractable feature representations. This bias causes the learning process of weaker modalities to lag behind, thereby inducing severe gradient imbalance. The compounding effect of this feature distribution deviation and the inconsistency in convergence behavior across modalities ultimately leads to numerical instability, or even training collapse, during the multi-modal fusion stage.

%To address the aforementioned challenges in multi modal collaborative optimization, this study analyzes the specific characteristics of hand drawn sketch, text, and image modalities to propose a staged training mechanism. This mechanism follows the primary principle of "training the subjects while freezing the background." By decoupling the optimization of network parameters, this study ensures a robust alignment of heterogeneous feature spaces throughout the training process.

To address the challenges, this study analyzes the characteristics of tri-modalities, and proposes a multi-stage cross-modal feature alignment mechanism. By decoupling the optimization trajectories of distinct modal encoders, our approach ensures robust alignment across multi-modal feature spaces.

\subsubsection{Sketch Feature Mapping.}
Considering the robust feature alignment already established between the image and text encoders of CLIP during the pretraining phase, MCFA prioritizes mapping the sketch feature space into this pre-aligned image-text space during the initial stage. To achieve this, the parameters of both the image and text branches within CLIP are strictly frozen, rendering the hand-drawn sketch encoder the sole optimizable component. the CLDRE and CKFSO modules are applied exclusively to the sketch modality. During this stage, the feature evolution of the three modalities is illustrated in \cref{eq:sfm}:

\begin{equation}
	\label{eq:sfm}
	\hat{f_S} = \mathtt{CLDRE}(f_S), \qquad \hat{f_T} = f_T, \qquad \hat{F_I} = F_I 
\end{equation}

\subsubsection{Image Feature Space Refinement.}
Recognizing that image and sketch modalities share greater visual affinity than text, MCFA proceeds to the second stage by fine-tuning the image encoder while keeping the sketch and text encoders frozen. This strategy leverages the geometric structural priors inherent in sketches to reshape the image feature distribution within the latent space, thereby enhancing the model's sensitivity to the topological layout of objects. The CLDRE and CKFSO modules are applied exclusively to the image modality. During this stage, the feature evolution of the three modalities is illustrated in \cref{eq:ifsr}:

\begin{equation}
	\label{eq:ifsr}
	\hat{f_S} = f_S, \qquad \hat{f_T} =f_T, \qquad \hat{F_I} = \mathtt{CLDRE}(F_I) 
\end{equation}

\subsubsection{Textual Representations Integration.}
Upon completing the complementary optimization of the sketch and image branches, we fine-tune the text encoder to ensure that fine-grained attributes (e.g., color, texture) are accurately embedded into the established joint feature space. In this stage, the image and sketch encoders are frozen, dedicating the optimization solely to enhancing the discriminative power of the textual representations. The CLDRE and CKFSO modules are applied exclusively to the text modality. During this stage, the feature evolution of the three modalities is illustrated in \cref{eq:tri}:

\begin{equation}
	\label{eq:tri}
	\hat{f_S} = f_S, \qquad \hat{f_T} =\mathtt{CLDRE}(f_T), \qquad \hat{F_I} = F_I 
\end{equation}

%The proposed staged training mechanism not only effectively mitigates the parameter divergence of multi modal networks during joint training at the underlying optimization level, but also logically simulates a cognitive process ranging from geometric framework construction to semantic attribute injection. This approach ensures a high degree of unity among the sketch, image, and text modalities within the final feature space, providing a robust representational foundation for achieving precise cross modal instance retrieval.

\subsection{Optimization Objectives}

Hand-drawn sketch features $\hat{f_S}$ and text features $\hat{f_T}$ are fused via element-wise addition to construct a composite query representation $\hat{f_c}$. Subsequently, cosine similarity is computed between the composite representation $\hat{f_c}$ and the candidate image features $\hat{F_I}$ within the gallery to yield the retrieval results.

The overall optimization objective is defined as follows:

\begin{equation}
	\label{eq:stbir_all_loss}
	L_{total} = \lambda_1 L_{AAML} + \lambda_2 L_{InfoNCE} + \lambda_3 L_{Triplet}
\end{equation}
where $L_{InfoNCE}$ and $L_{Triplet}$ represent the InfoNce loss and Triplet loss widely adopted in retrieval tasks, respectively, which take the composite representation $\hat{f_c}$ and candidate image features $\hat{F_I}$ as inputs, aiming to maximize feature discriminability. $\lambda_1$, $\lambda_2$, and $\lambda_3$ serve as the balancing weight coefficients for the respective loss terms.

\section{Expriment}
\subsection{Dataset}
The experimental validation of this research is primarily conducted on the STBIR dataset introduced in \cref{sec:dataset}.
\subsection{Training Details}
All experiments are deployed on a single NVIDIA RTX 3090 GPU. Regarding network initialization, the hand-drawn sketch branch utilizes ResNet50 \cite{he2016deep} as the feature encoder, which is pretrained on the QuickDraw dataset\cite{ha2017neural}. The image and text encoders directly load the pretrained weights from CLIP \cite{radford2021learning}. In the MCFA, each stage is trained for 16 epochs. AdamW is used as the optimizer, with the learning rate set to $2 \times 10^{-5}$ and the batch size configured at 32. For \cref{eq:stbir_arc_face}, $s$ and $m$ are set to 32 and 0.15. For \cref{eq:stbir_all_loss}, $\lambda_1$, $\lambda_2$ and $\lambda_3$ are set to 0.1, 0.8 and 0.8.

\subsection{Metrics}
In alignment with the quantitative evaluation of retrieval performance, this research references classical works in the retrieval field \cite{sangkloy2022sketch,koley2024you} and adopts the standard retrieval metric Recall@K (R@K) as the primary evaluation criterion, specifically covering R@1, R@5, and R@10. The R@K metric measures the proportion of samples where the ground truth target is included within the top K candidate samples retrieved by the model.

\subsection{Benchmark}
The evaluated baseline methods comprise two categories: single query retrieval and multi query retrieval. The former leverages classic dual stream networks using a single query modality. The latter includes models specifically designed for sketch and text based image retrieval, alongside adapted multimodal approaches.

\subsection{Performance Analysis}
As evidenced by the experimental results in \cref{tab:stbir_comparison_centered}, the proposed STBIR framework achieves superior retrieval performance across all three datasets, outperforming baseline methods in nearly all metrics.

\begin{table}[t]
	\centering
	\footnotesize 
	\caption{Performance comparison on the STBIR-S, STBIR-C, and STBIR-D datasets. \textbf{Bold} values indicate the best results. \underline{Underlined} values indicate the second-best results.}
	\label{tab:stbir_comparison_centered}
	% 在第一列(l)和第二列(l)之间插入 @{\hspace{2em}} 来显著增加间距
	\begin{tabularx}{\textwidth}{l @{\hspace{1em}} l *{9}{>{\centering\arraybackslash}X}}
		\toprule
		& & \multicolumn{3}{c}{STBIR-S} & \multicolumn{3}{c}{STBIR-C} & \multicolumn{3}{c}{STBIR-D} \\
		\cmidrule(lr){3-5} \cmidrule(lr){6-8} \cmidrule(lr){9-11}
		Query & Method & R@1 & R@5 & R@10 & R@1 & R@5 & R@10 & R@1 & R@5 & R@10 \\ \midrule
		
		\multirow{3}{*}[0pt]{\makecell{Single\\Query}} 
		& CLIP(S)\cite{radford2021learning}      & 2.40  & 10.06 & 17.12 & 10.29 & 26.05 & 38.59 & 5.05  & 17.03 & 24.50 \\
		& CLIP(T)\cite{radford2021learning}      & 46.00 & 81.50 & 92.50 & 48.00 & 82.00 & 92.00 & 53.57 & 87.89 & 96.63 \\
		& DINO(S)\cite{simeoni2025dinov3}      & 18.17 & 38.74 & 51.50 & 48.23 & 76.85 & 84.24 & 26.41 & 61.74 & 78.15 \\ \midrule
		
		\multirow{4}{*}[0pt]{\makecell{Multi\\Query}} 
		& CLIP(ST)\cite{radford2021learning}     & 22.97 & 52.55 & 67.57 & 41.80 & 72.03 & 83.28 & 27.24 & 62.96 & 77.32 \\
		& TASKformer\cite{sangkloy2022sketch}  & 39.49 & 72.07 & 83.78 & \underline{53.38} & 84.89 & 92.93 & 51.70 & 79.83 & 88.03 \\
		& Pic2Word\cite{saito2023pic2word}    & \underline{51.72} & 82.58 & 89.78 & \underline{53.38} & 86.81 & 95.17 & 51.33 & 85.08 & 94.43 \\
		& SEARLE\cite{baldrati2023zero}      & 49.70 & \underline{87.39} & \underline{94.74} & 52.41 & \underline{87.46} & \underline{95.18} & \underline{62.43} & \underline{92.36} & \textbf{98.38} \\ \midrule
		
		& STBIR  & \textbf{51.80} & \textbf{87.84} & \textbf{96.70} & \textbf{57.88} & \textbf{92.28} & \textbf{97.75} & \textbf{62.85} & \textbf{93.44} & \underline{98.06} \\
		\bottomrule
	\end{tabularx}
\end{table}

\subsubsection{Single-Query vs. Multi-Query Settings.}
Experimental results indicate that multi-query retrieval architectures generally outperform single-query setups. Purely sketch-based approaches, however, yield inferior performance, as the lack of texture and color information impedes fine-grained retrieval. In contrast, CLIP(T) leverages priors from large-scale text-image pretraining to achieve robust semantic alignment. Notably, the proposed STBIR framework significantly enhances retrieval performance by effectively fusing multimodal query information, thereby underscoring the critical value of synergizing stroke structures with textual semantics.

\subsubsection{Comparison with State of the Art Methods. }
Experimental results demonstrate that the proposed STBIR framework consistently outperforms existing state of the art multi-query retrieval methods across all three benchmark datasets. Specifically, STBIR achieves superior performance on the STBIR-S and STBIR-C datasets across all evaluation metrics. Notably, on the STBIR-C dataset, it attains an R@1 accuracy of 57.88, significantly surpassing the second-best models, TASKformer and Pic2Word. On the larger-scale STBIR-D dataset, while STBIR trails the SEARLE model slightly in R@10, it secures the top rank on the more critical R@1 and R@5 metrics, achieving 62.85 and 93.44, respectively. These substantial gains underscore the effectiveness of our method in fusing text and sketch modalities, optimizing the multimodal feature space, and enhancing instance-level discriminability.

\subsubsection{Robustness Analysis in Many-Category Scenarios.}
On the STBIR-D dataset, which presents a more challenging category scale, the proposed framework maintains state of the art performance in both R@1 (62.85) and R@5 (93.44). While SEARLE \cite{baldrati2023zero} achieves a marginal gain on the R@10 metric, the STBIR framework demonstrates a substantial advantage on the R@1 metric, where precision requirements are most stringent. This performance underscores the robust capability of STBIR to effectively optimize the feature space, even when confronted with large-scale category distributions.

\subsection{Ablation Study}
As illustrated in \cref{tab:stbir_ablation_comparison}, the effectiveness of our framework is validated across four key aspects input modalities, training mechanism, optimization objective, and robustness enhancement strategies. 

\begin{table}[t]
	\centering
	\footnotesize 
	\caption{Ablation study results on the STBIR-S and STBIR-C datasets. The 'Te' and 'Sh' denote the text and sketch modalities, respectively, indicating that the model inputs are standard raw text and sketch data, rather than feature vectors processed via global mean pooling. \textbf{Bold} values indicate the best results.}
	\label{tab:stbir_ablation_comparison}
	% 前5列放置对勾，使用 c 保持紧凑；后6列放置数据，使用 X 平分剩余的文本宽度
	\begin{tabularx}{\textwidth}{ccccc *{6}{>{\centering\arraybackslash}X}}
		\toprule
		\multicolumn{5}{c}{Ablation Setting} & \multicolumn{3}{c}{STBIR-S} & \multicolumn{3}{c}{STBIR-C} \\
		\cmidrule(lr){1-5} \cmidrule(lr){6-8} \cmidrule(lr){9-11}
		Te & Sh & MCFA & CKFSO & CLDRE & R@1 & R@5 & R@10 & R@1 & R@5 & R@10 \\ \midrule
		
		& \checkmark & \checkmark & \checkmark & \checkmark & 4.50 & 13.96 & 21.02 & 6.43 & 20.90 & 35.37 \\
		\checkmark &  & \checkmark & \checkmark & \checkmark & 41.74 & 78.98 & 88.74 & 37.94 & 79.74 & 91.32 \\
		\checkmark & \checkmark &  & \checkmark & \checkmark & 7.81 & 41.89 & 62.31 & 28.62 & 67.85 & 84.24 \\
		\checkmark & \checkmark & \checkmark &  & \checkmark & 47.75 & 84.98 & 93.99 & 55.95 & \textbf{92.28} & 97.11 \\
		\checkmark & \checkmark & \checkmark & \checkmark &  & 47.85 & 84.68 & 94.29 & 43.73 & 78.78 & 88.75 \\ \midrule
		
		\checkmark & \checkmark & \checkmark & \checkmark & \checkmark &\textbf{51.80} & \textbf{87.84} & \textbf{96.70} & \textbf{57.88} & \textbf{92.28} & \textbf{97.75} \\
		\bottomrule
	\end{tabularx}
\end{table}

Experimental results indicate that each core component proposed in this study is indispensable for the collaborative enhancement of retrieval performance. First, the complementarity of composite modal inputs is strongly validated (Te and Sh in \cref{tab:stbir_ablation_comparison}); the absence of any single modality leads to a significant degradation in representational capability. Second, the multi-stage cross-modal feature alignment mechanism proves critically important (MCFA in \cref{tab:stbir_ablation_comparison}). Adopting synchronous updates for all parameters causes a sharp decline in the R@1 metric, demonstrating that full-parameter fine-tuning can easily induce semantic drift in pretrained weights and trigger inter-modal feature interference. Furthermore, both the category-knowledge-based feature space optimization module (CKFSO in \cref{tab:stbir_ablation_comparison}) and the curriculum learning driven robustness enhancement module (CLDRE in \cref{tab:stbir_ablation_comparison}) contribute substantially to achieving high-precision, fine-grained retrieval.

\subsection{Multi-stage Cross-modal Feature Alignment Mechanism Analysis}
This section further explores the impact of different modality training sequences within the MCFA, with experimental results presented in \cref{tab:stbir_optimization_order_expanded}. Experimental data indicate that training schemes prioritizing the optimization of sketch features (as shown in the last two rows of \cref{tab:stbir_optimization_order_expanded}) yield superior retrieval performance. This phenomenon reflects that in composite retrieval tasks, the sketch modality possesses extreme geometric abstraction and lacks a foundation of large scale joint pretraining with other modalities; therefore, it is essential to prioritize the establishment of a robust feature mapping benchmark during the initial training phase.

\begin{table}[t]
	\centering
	\footnotesize 
	\caption{Comparison of different modality training sequences on retrieval performance. The I, S, and T represent the Image, Sketch, and Text modalities. The arrow notation (e.g., $\text{I} \to \text{S} \to \text{T}$ ) specifies the sequential optimization schedule of the corresponding encoders. \textbf{Bold} values indicate the best results.}
	\label{tab:stbir_optimization_order_expanded}
	% 第一列权重增加至 2.0，其余 6 列平分剩余空间（权重约为 0.83）
	\begin{tabularx}{\textwidth}{>{\hsize=2.0\hsize\centering\arraybackslash}X *{6}{>{\hsize=0.83\hsize\centering\arraybackslash}X}}
		\toprule
		& \multicolumn{3}{c}{STBIR-S} & \multicolumn{3}{c}{STBIR-C} \\
		\cmidrule(lr){2-4} \cmidrule(lr){5-7}
		Optimization Order & R@1 & R@5 & R@10 & R@1 & R@5 & R@10 \\ \midrule
		
		$\text{I} \to \text{S} \to \text{T}$ & 47.15 & 84.83 & 94.74 & 53.38 & 88.42 & 95.50 \\
		$\text{I} \to \text{T} \to \text{S}$ & 47.45 & 87.69 & 94.14 & 46.95 & 89.07 & 95.50 \\
		$\text{T} \to \text{S} \to \text{I}$ & 47.15 & 84.98 & 94.14 & 51.77 & 81.67 & 90.03 \\
		$\text{T} \to \text{I} \to \text{S}$ & 47.75 & 83.93 & 93.24 & 54.98 & 84.24 & 91.32 \\ 
		$\text{S} \to \text{T} \to \text{I}$ & 48.05 & 84.38 & 94.59 & 57.23 & 89.07 & 96.46 \\ \midrule
		
		$\text{S} \to \text{I} \to \text{T}$ & \textbf{51.80} & \textbf{87.84} & \textbf{96.70} & \textbf{57.88} & \textbf{92.28} & \textbf{97.75} \\
		\bottomrule
	\end{tabularx}
\end{table}

Furthermore, given a sketch-first strategy, optimizing the image branch prior to the text branch yields superior retrieval performance. This advantage stems from the fact that although hand-drawn sketches and natural images exhibit significant domain shifts, they both belong to the visual domain and share strong correlations in spatial structure, contours, and geometric topology.  In contrast, text represents high-level abstract semantics, creating a substantially larger modality gap with sketches. Therefore, adopting an evolutionary progression from visual to semantic information more effectively guides the model toward deep cross-modal alignment.

\subsection{Visualization Analysis}

\begin{figure}[t] % h:当前位置, t:顶部, b:底部, p:独立一页
	\centering
	\includegraphics[width=\textwidth]{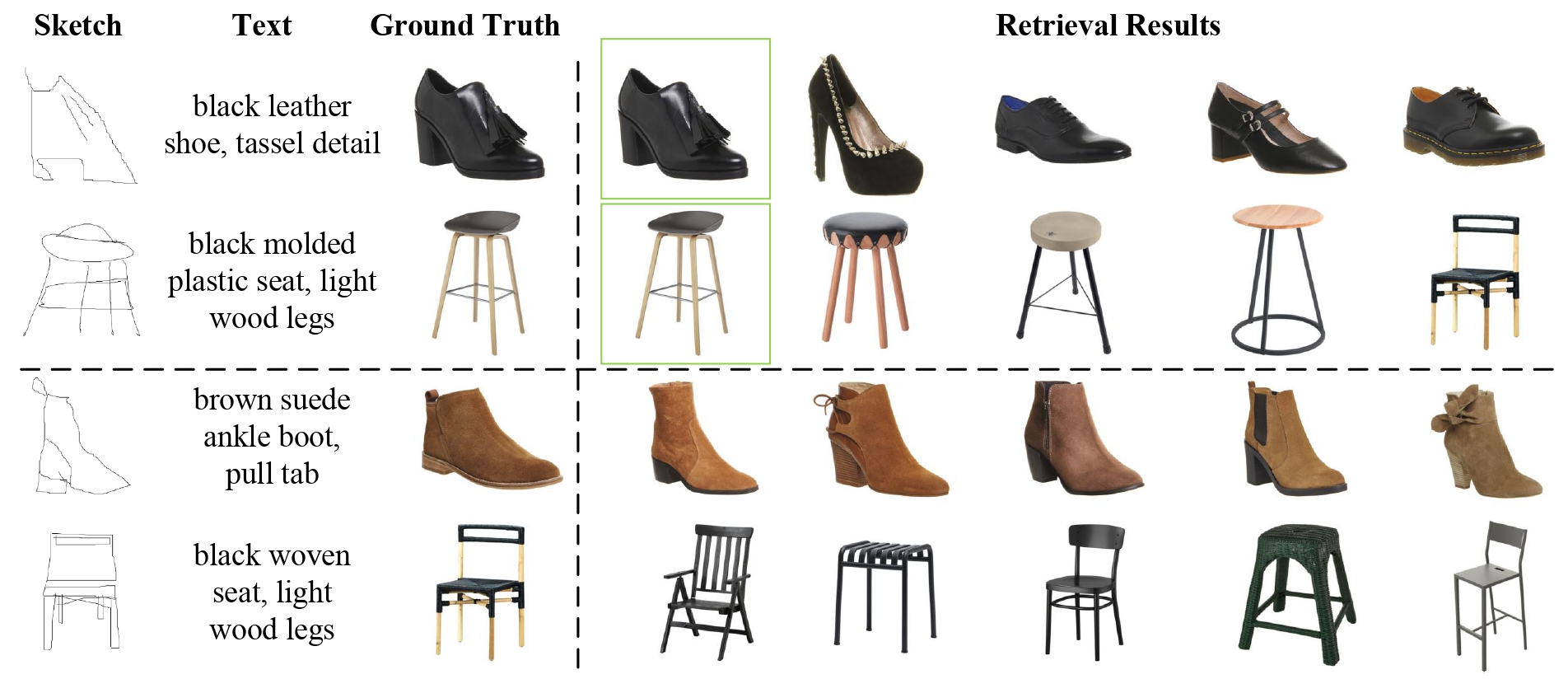} 
	\caption{Visualization of some retrieval results. Retrieved candidates are ranked in descending order based on their predicted scores. Instances enclosed in green boxes denote correctly retrieved samples.} % 用于文中引用，如：见图 \ref{fig:my_label}
	\label{fig:visual}
\end{figure}

\cref{fig:visual} visualizes representative retrieval results: the first two rows showcase successful retrievals (where the ground truth ranks within the top-5), while the last two rows illustrate failure cases. The figure reveals that nearly all retrieved samples exhibit high consistency with the query conditions (sketch + text) in terms of contour, texture, and color, underscoring STBIR's robust capability for multi-modal feature alignment.

Notably, even the erroneously retrieved samples in the last two rows adhere strictly to the provided sketch and textual descriptions. This suggests that the bottleneck in discriminating highly similar instances lies not only in the model's capacity to capture fine-grained local features, but more critically, in the inherent ambiguity of hand-drawn sketches and the limited granularity of textual descriptions at the input stage. This observation highlights a pivotal direction for future research aimed at enhancing composite retrieval performance.

\section{Conclusion}
In this paper, we propose a comprehensive and robust framework to advance the state of the art in sketch and text based image retrieval. First, we construct and release the STBIR dataset, which comprises the STBIR-C and STBIR-S subsets designed for fine-grained retrieval within single categories, as well as the STBIR-D dataset tailored for large-scale category retrieval scenarios. Second, we introduce a category-knowledge-based feature space optimization module and a curriculum learning-driven robustness enhancement module to jointly enhance the model's feature discriminability and retrieval robustness. Furthermore, we propose a multi-stage cross-modal feature alignment mechanism grounded in the distinct characteristics of each modality, thereby achieving alignment across multi-modal feature spaces. By integrating these modules, the proposed STBIR framework significantly optimizes the alignment across sketch, text, and image feature spaces, ultimately achieving superior retrieval performance.

\bibliographystyle{splncs04}
\bibliography{main}
\end{document}